\documentclass{sigkddExp}

\DeclareMathOperator*{\argmax}{arg\,max}
\DeclareMathOperator*{\argmin}{arg\,min}
\newtheorem{definition}{Definition}
\usepackage{tabularx}
\usepackage{multirow}
\usepackage{booktabs}
\usepackage{tikz-qtree}
\usepackage{tikz-qtree-compat}
\usepackage{hyperref}

\usepackage{color}
\usepackage{tikz}
\usepackage{soul}
\usepackage{amsmath}

\newcommand{\EX}{\operatorname{\mathbb{E}}}

\usetikzlibrary{arrows,shapes,positioning,shadows,trees}

\tikzset{
  basic/.style  = {draw, text width=2cm, drop shadow, font=\sffamily, rectangle},
  root/.style   = {basic, rounded corners=2pt, thick, align=center,
                   fill=white!30, text width=15em, text height=1.5em},
  level 2/.style = {basic, rounded corners=2pt, thick,align=center, fill=white!30,
                   text width=10em, text height=1.5em},
  level 3/.style = {basic, thick, align=left, fill=white!30, text width=8em}
}

\begin{document}
%
% --- Author Metadata here ---
% -- Can be completely blank or contain 'commented' information like this...
%\conferenceinfo{WOODSTOCK}{'97 El Paso, Texas USA} % If you happen to know the conference location etc.
%\CopyrightYear{2001} % Allows a non-default  copyright year  to be 'entered' - IF NEED BE.
%\crdata{0-12345-67-8/90/01}  % Allows non-default copyright data to be 'entered' - IF NEED BE.
% --- End of author Metadata ---

\title{Causal Interpretability for Machine Learning \\ - Problems, Methods and Evaluation}

\author{Raha Moraffah$^*$, Mansooreh Karami$^*$, Ruocheng Guo$^*$, Adrienne Raglin$^{\dagger}$, Huan Liu$^*$\\\affaddr{$^*$Computer Science \& Engineering, Arizona State University, Tempe, AZ, USA}\\\affaddr{$^{\dagger}$Army Research Lab, USA}\\\affaddr{$^*$\{rmoraffa, mkarami, rguo12, huanliu\}@asu.edu, $^{\dagger}$adrienne.raglin2.civ@mail.mil} }
%\subtitle{[Extended Abstract]
% You need the command \numberofauthors to handle the "boxing"
% and alignment of the authors under the title, and to add
% a section for authors number 4 through n.
%
% Up to the first three authors are aligned under the title;
% use the \alignauthor commands below to handle those names
% and affiliations. Add names, affiliations, addresses for
% additional authors as the argument to \additionalauthors;
% these will be set for you without further effort on your
% part as the last section in the body of your article BEFORE
% References or any Appendices.

%\numberofauthors{5}
%
% You can go ahead and credit authors number 4+ here;
% their names will appear in a section called
% "Additional Authors" just before the Appendices
% (if there are any) or Bibliography (if there
% aren't)

% Put no more than the first THREE authors in the \author command
%%You are free to format the authors in alternate ways if you have more 
%%than three authors.

\title{Causal Interpretability for Machine Learning \\ - Problems, Methods and Evaluation}

\maketitle

\begin{abstract}
Machine learning models have had discernible achievements in a myriad of applications. However, most of these models are black-boxes, and it is obscure how the decisions are made by them. This makes the models unreliable and untrustworthy. To provide insights into the decision making processes of these models, a variety of traditional interpretable models have been proposed. Moreover, to generate more human-friendly explanations, recent work on interpretability tries to answer questions related to causality such as ``Why does this model makes such decisions?'' or ``Was it a specific feature that caused the decision made by the model?''. In this work, models that aim to answer causal questions are referred to as causal interpretable models.
The existing surveys have covered concepts and methodologies of traditional interpretability.
In this work, we present a comprehensive survey on causal interpretable models from the aspects of the problems and methods. In addition, this survey provides in-depth insights into the existing evaluation metrics for measuring interpretability, which can help practitioners understand for what scenarios each evaluation metric is suitable.
%make sense of what machine learning frameworks are doing and how they are making their decisions, several interpretable models have been proposed.
\end{abstract}



%
% Keywords. The author(s) should pick words that accurately describe the work being
% presented. Separate the keywords with commas.
\keywords{Interpratablity, explainability, causal inference, counterfactuals, machine learning} %neural networks, }

%
% A "teaser" image appears between the author and affiliation information and the body 
% of the document, and typically spans the page. 

%
% This command processes the author and affiliation and title information and builds
% the first part of the formatted document.
\maketitle

\section{Introduction}
%What is interpretable machine learning and why is it necessary to have it?
With the surge of machine learning in critical areas such as healthcare, law-making and autonomous cars, decisions that had been previously made by humans are now made automatically using these algorithms. 
%
% In order for humans to be able to trust such decisions
In order to ensure the reliability of such decisions, humans need to understand how these decisions are made.
%\rg{I think interpretability should mean more than reliability, for example, the two following examples are more about fairness instead of reliability. In fact there are three aspects of motivation covered by this paragraph: interpretability itself, imposing fairness and defending adversarial attacks.}
%
However, machine learning models are usually inherently black-boxes and do not provide explanations for \emph{how} and \emph{why} they make such decisions. This has become especially problematic when recent work shows that the decisions made by machine learning models are sometimes biased and enforce inequality \cite{mehrabi2019survey}.
For instance, Angwin et al. \cite{angwin_larson_kirchner_mattu_2019} demonstrates that predictions made by Correctional Offender Management Profiling for Alternative Sanctions (COMPAS), which is a widely used criminal risk assessment tool, shows racial biases. %\raha{we can talk about google classifier incident as well}.
%\rg{To show how interpretable models help solve such problems, for the example, maybe you can say something like if we have an interpretable model, we can avoid such unfairness with human intervention.}
%
With recent regulations such as European Union’s ``Right to Explanation'' \cite{goodman2017european} and AI call for diversity and inclusion \cite{boyd2012critical}, interpretable models which are capable of explaining the decisions they made are necessary.
%the need for interpretable \footnote{In this survey we use the words interpretable and explainable interchangeably.} models which are capable of explaining the decisions they made is felt.
%\rg{I think this example should be the first because the fact that humans aim to understand the decision making of machines itself is a motivation and then interpretability also helps us to solve the problems as unfairness and adversarial attacks.}
%
Moreover, recent research shows that machine learning models, especially deep neural networks, can be easily fooled into predicting a specific class label for an image when its pixel values are under minimal perturbations \cite{goodfellow2014explaining, DBLP:journals/corr/Moosavi-Dezfooli15,DBLP:journals/corr/PapernotMJFCS15}.
Such results imply that machine learning models suffer from the risk of making unexpected decisions.
Understanding decisions of machine learning models and the process leading to decision making can help us understand the rules the models use to make their decisions and therefore, prevent potential unexpected situations from happening. 
%\rg{Could you add a sentence to clarify: if we have interpretable models, how do they avoid being fooled by adversarial examples?}
%
More specifically, through interpretable machine learning models, we aim to guarantee that (a) decisions made by machine learning models comply with the rules toward social good;
(b) the classifier does not pick up the biases in the data and the decisions made are compatible with human understandings.

Previously, various frameworks have been proposed to generate explanations for machine learning algorithms.
These algorithms can be mainly divided into two categories, (1) algorithms that are inherently interpretable, which includes the models that generate explanations at training time \cite{yang2016hierarchical}; (2) post-hoc interpretations that refer to the model that generate explanations for already made decisions \cite{Mordvintsev45507,plumb2018model, kim2016examples}. Henceforth, these models are referred to as traditional interpretable models.
%example based explanations, which mainly focus on finding examples in the data that can explain the decisions the best \cite{kim2016examples}  and (2) model based explanations that aim to explain the effect of different components of a classifier on the decisions \cite{koh2017understanding,Mordvintsev45507,plumb2018model}.% \mk{\st{made}}  \raha{why causal explanations} 
%

% \rg{You may want to say that the traditional methods cannot answer these question and explain why they cannot briefly.}

%\rg{To motivate the need for causal interpretability, can you explain the why answering these questions are necessary? (Compared to traditional interpretability, how answering these questions help human beings understand more about the models? What are the problems traditional methods cannot solve but causal interpretable models can?)} \raha{one or two examples of why causal based interpretability and answering these questions are useful}
%
% While looking for explanations, we look for the answers which consider alternative solutions and situations.
In this work, we focus on causal interpretable models that can explain their decisions through what decisions would have been made if they had been under alternative situations (e.g., being trained with different inputs, model components or hyperparameters).
%
%\rg{Maybe you could add a typical question that can be answered by traditional interpretable models as a comparison. This comparison may also help show that only with causal interpretable models some problems can be solved such as "how can a person improve her profile to get application accepted?".} \raha{influence function}
%
Note that traditional interpretable models are unable to answer such questions about decision making under alternative situations, although they can explain how and why a decision is made by an existing model on an observed instance. 
% Traditional interpretable models can be used to measure the effect of a data sample on final decision \cite{koh2017understanding}. 
%
% However, they are unable to answer questions about alternative feature assignment or situations.
%
For instance, in the case of credit applications, to impose fairness on the decision making process,
we may need to answer questions such as Did the protected features (e.g., race and gender etc.) cause the system to reject the application of the $i$-th applicant?'' and ``If the $i$-th applicant had different protected features, would the system still make the same decision?''
%\rg{As it seems we want to focus on addressing the fairness issue through causal interpretability in the examples, I modified them a little. In addition, to avoid conflicts in notations, I suggest to remove the $X,Y,Z$ used here as they are likely to be used later.}
%"why did the self-driving car classify the stop sign as fire hydrant?"
%
In other words, in order to make the explanations more understandable and useful for humans, we need to ask questions such as ``Why did the classifier make this decision instead of another?'', ``What would have happened to this decision of a classifier had we had a different input to it?'', or ``Was it feature $X$ that caused decision $Y$?''.
%\rg{does X mean a certain feature or a model component or something else? also try to avoid using notations here to avoid conflicts.}
%
Traditional interpretability frameworks which only consider correlations are not capable of generating such explanations. This is due to the fact that these frameworks cannot estimate how altering a feature or a component of a model would change the predictions made by the rest of the model or the predicted labels on the data samples. % \cite{}.
Therefore, in order to answer such questions about both data samples and models, counterfactual analysis needs to be leveraged. Counterfactual analysis is a concept from the causal inference literature \cite{article}. In counterfactual analysis, we aim to infer the output of a model in imaginary scenarios that we have not observed or cannot observe.
Recently, counterfactual analysis and causal inference have gained a lot of attention from the interpretable machine learning field. Research in this area has mainly focused on generating counterfactual explanations from both the data perspective \cite{DBLP:journals/corr/abs-1904-07451,DBLP:journals/corr/abs-1905-07697} as well as the components of a model \cite{DBLP:journals/corr/abs-1811-04376,DBLP:journals/corr/abs-1802-00541}. 

%\rg{Compared to existing surveys about interpretability, what is new and why this survey is important?}
%\raha{one sentance introduction of existing surveys and comparing this one with them}
Existing surveys on interpretable machine learning focus on the traditional methods and do not discuss the existing methods from a causal perspective. In this survey, we present commonly used definitions for interpretability, discuss interpretable models from a causal perspective and provide guidelines for evaluating these methods.
More specifically, in Section \ref{def}, we first provide different definitions for interpretability. We then briefly introduce the existing methods on traditional interpretablity and present different types of interpretable models in this category (Section \ref{traditional}). Section \ref{causal} discusses concepts from causal inference, which are used in this survey. In section \ref{causalinterpret}, we provide an overview of existing works on causal interpretability. We also compare the proposed models for both traditional and causal models from different perspectives to provide insights on advantages and disadvantages of each type of interpretability. Section \ref{eval} provides detailed guidelines on the experimental settings such as commonly used datasets and evaluation metrics for both traditional and causal approaches. We then discuss evaluation metrics specifically used for causal methods in more detail and provide different scenarios for which these metrics can be used. Since the evaluation of causal interpretable models is a challenging task, these guidelines can be helpful for future research in this area and can be used to evaluate approaches with similar characteristics. In addition, they can also be used to create new evaluation metrics for the approaches with different functionalities.

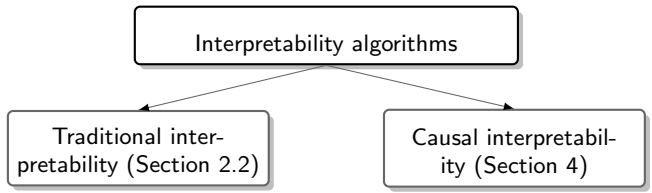
\begin{figure}[htp!]
\begin{tikzpicture}[
  level 1/.style={sibling distance=50mm, draw=black!60},
  edge from parent/.style={->,draw},
  >=latex]

% root of the the inInterl 1
\node[root] {Interpretability algorithms}
% The first level, as children of the initial tree
  child {node[level 2, text height=1em] (c1) {Traditional interpretability (Section \ref{traditional})}}
  child {node[level 2, text height=1.1em] (c2) {Causal interpretability (Section \ref{causalinterpret})}};
  %child {node[level 2] (c3) {Drawing arrows between nodes}};

% The second level, relatively positioned nodes

% lines from each level 1 node to every one of its "children"
%\foreach \value in {1,2}
%\draw[->] (c1) |- (c2);

%\foreach \value in {1,2,3}
%  \draw[->] (c2.195) |- (c2\value.west);

\end{tikzpicture}

\caption{Main categories for Interpretable frameworks} %\rg{as causal interpretability appears in this figure, the categories should also be briefly introduced in this section.}}
\label{fig:outline}
\end{figure}

\section{An overview of Interpretability} \label{def}
%Here we will talk about different definitions of interepretability and its categories(briefly)
In this section, we present an overview of existing definitions for interpretability.
%
%Then, we categorize the existing frameworks based on these definitions. %We will also briefly discuss the features the an interpretable model should have in order to be able to generate good and useful explanations.
Miller et al. \cite{DBLP:journals/corr/Miller17a} suggest that interpretability is the degree to which a human can understand the cause of a decision.
Kim et al. \cite{DBLP:conf/nips/KimKK16} propose that interpretability is the degree to which a human can consistently predict the model's decisions.
Doshi-Velez et al. \cite{doshi2017towards} define interpretability as the ability to explain in intelligible ways to a human.
Gilpin et al. \cite{gilpin2018explaining} take a step further and define interpretability as a part of explainability. They state that explainable models are those that summarize the reasons for neural network behaviors, gain the trust of the users, or generate insights into the causes of their decisions while interpretable models may not be able to describe the operation of a system in an accurate way \footnote{In this survey we use the words interpretable and explainable interchangeably.}. Pearl \cite{Pearl:2019:STC:3314328.3241036} claims that tasks such as explainability
% a sub-task of robustness, 
require a causal model of the environment and cannot be handled at the level of association. 

\subsection{Interpetability in Machine Learning}
%\raha{later we make it only a graph}
Interpretable machine learning has been widely explored and discussed in previous literature. However, to the best of our knowledge, there is no comprehensive review on causal interpretability models. For instance, Lipton \cite{DBLP:journals/corr/Lipton16a} discusses the motivation behind creating interpretable models and categorizes interpretable models into two main categories: transparent models and post-hocs. Doshi-velez et al. \cite{doshi2017towards} provide a definition of model interpretability and evaluation criteria. However, this review only proposes definitions and evaluations that are used for traditional interpretability of models and does not cover causal and counterfactual questions.
Gilpin et al. \cite{gilpin2018explaining}, explain fundamental concepts of explainability and use them to classify the literature on interpretable models. Zhang and Zhu \cite{zhang2018visual} review the existing interpretable models proposed for deep models used in visual domains. Du et al. \cite{du2018techniques} provide a comprehensive survey of existing interpretable methods and discuss issues that should be considered in future work. It is worth mentioning that none of the existing work discussed interpretable models from a causal perspective.
In this work, we first introduce the state-of-the-art research in traditional  interpretability (Sec. \ref{traditional}) and then give a detailed survey on causal interpretable models (Sec. \ref{causalinterpret}).
Figure \ref{fig:outline} shows an overview of intepretable models and their classification.
%In what follows, we will give a quick introduction to traditional interpretable frameworks, classify them into two main categories and discuss state-of-the-art approaches for each category. The two main categories of the existing works on traditional interpretable machine learning are:

%
%We first give an overview of inherently and post-hoc interpretable methods in traditional interpretability. 
%
%Then, we provide a brief introduction of causal inference concepts which are necessary for understanding the causal based interpretability.
%
%Finally, a comprehensive review of existing approaches on causal interpretablity is presented. %can be categorized into \textit{model-based} and \textit{example-based} interpretation and will be explored in details.
%Here we give a formal definition of interpretability
\subsection{Traditional Interpretablity} \label{traditional}

Before proceeding with the detailed review of the methodologies in causal interpretable models, we provide an overview of existing state-of-the-art methods in traditional machine learning. We categorize traditional models into two main categories:

\begin{itemize}
    %\item Example-based interpretablity: Which is looking for examples from the dataset which explain the model's behavior the best.
    \item Inherently interpretable models: Models that generate explanations in the process of decision making or while being trained.
    \item Post-hoc interpretability: Generating explanations for an already existing model using an auxiliary model. Example-based interpretablity also falls into this category. In example-based interpretablity, we are looking for examples from the dataset which explain the model's behavior the best.
\end{itemize}
%we present in this section the largely used interpretable approaches in machine learning. We will refer to these as traditional interpretable machine learning models.

% \mk{Recent supervised learning algorithms heavily depend on the data to extract informative knowledge. The first step to create interpretable machine learning algorithms is to understand the useful features and bold characteristics of the data. There are models in which the features are meaningful and usually are perceptible for humans, while others might include features that are incomprehensible to humans \cite{ilyas2019adversarial}. 
% \\Interpretability for data does not necessarily means that for every model each individual feature or sample should be defined by text description, but it should admits an intuitive explanation \cite{DBLP:journals/corr/Lipton16a}.
% \\}

\subsubsection{Interpretable Models} \label{inr_clssi}
A machine learning model can be designed to include explanations embedded as part of their architecture or output interpretable decisions as part of their training process. Most of these models are created in application of the deep neural network.
In this section, we present common interpretable models in the literature.

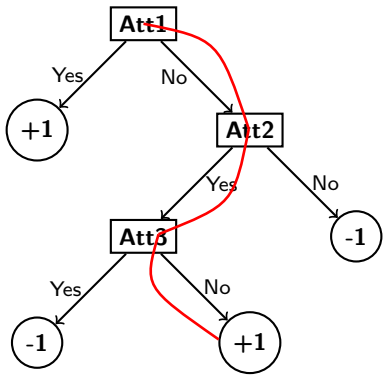
\begin{figure}[h]
\centering
        \begin{tikzpicture}[->, node distance=2cm, every loop/.style={},
                            thick,leaf node/.style={circle,draw,font=\sffamily\Large\bfseries}, main node/.style={draw,font=\sffamily\Large\bfseries}]
        
          \node[main node] (1) {\normalsize Att1};
          \node[leaf node] (2) [below left of=1] {\normalsize +1};
          \node[main node] (3) [below right of=1] {\normalsize Att2};
          \node[main node] (4) [below left of=3] {\normalsize Att3};
          \node[leaf node] (5) [below right of=3] {\normalsize -1};
          \node[leaf node] (6) [below left of=4] {\normalsize -1};
          \node[leaf node] (7) [below right of=4] {\normalsize +1};

          \path[every node/.style={font=\sffamily\small}]
                (1) edge node [left] {Yes} (2)
                (1) edge node [left] {No} (3)
                (3) edge node [right] {Yes} (4)
                (3) edge node [right] {No} (5)
                (4) edge node [left] {Yes} (6)
                (4) edge node [right] {No} (7);
        \draw[-,line width=1pt, color = red] (0,0) .. controls (1,-0.2) .. (1.4, -1.4);
        \draw[-,line width=1pt, color = red] (1.4, -1.4) .. controls (1.2,-2.4) .. (0.2, -2.8);
        \draw[-,line width=1pt, color = red] (0.2, -2.8) .. controls (0,-3.5) .. (1, -4.2);
        \end{tikzpicture}
  \caption{An example of a decision tree with positive and negative class (binary) and three attributes. The red path has a decision rule, if $\neg$\textit{Att1} $\wedge$ \textit{Att2} $\wedge$ $\neg$\textit{Att3} $\Rightarrow$ +1} 
  \label{fig:DT}
\end{figure}

\textbf{Decision Trees.} These methods make use of a tree-structured framework in which each internal node checks whether a condition on a feature is satisfied or not while the leaf nodes show the final predictions (class labels). 
A decision infers the label of an instance by starting from the root and tracing a path till a leaf node is reached, which can be interpreted as an \emph{if..then..} rule.
An example is illustrated in Figure \ref{fig:DT}.

\textbf{Rule-Based Models.} Rule-based classifiers also create explanations that are interpretable for humans. These classifiers use a collection of \textit{if..then..} rules to infer the class labels. In a sense, rule-based classifiers are the text representation of the decision trees. However, there are some key differences. Rule-based models can have rules that are not mutually exclusive (i.e., two or more rules might trigger by the same record), not exhausted (i.e., a record may not trigger any rules) and ordered (i.e., the rule set is ordered based on their priority) \cite{tan2018introduction}. 

\textbf{Linear Regression.} Another common method known to be interpretable is Linear Regression.
Linear Regression models the linear relation between a dependent variable and a set of explanatory variables (features).
The weight of each feature represents the mean change in the prediction given a one unit increase of the feature.
Accordingly, it is reasonable to think that the features with larger weights has more effect on the final result. 
However, different types of variables (e.g., categorical data vs numerical features) have different scales.
This makes it difficult to interpret the effect of each feature. Fortunately, there are several methods that can be used to find the importance of a feature in a linear regression such as t-statistics and chi-square score \cite{li2018feature}.

 The aforementioned methods are restricted by users' limitations (i.e., human understanding). With the increase in the number of features, these models become more and more complex;  for example, decision trees become much deeper, and the number of the rules increase in the rule sets. This makes comprehending the prediction of these models difficult for humans \cite{ribeiro2016should}. Below, we discuss recent inherently interpretable models which are designed for more sophisticated scenarios.

\textbf{Attention Networks.} Attention networks have been successful in various highly-impactful tasks such as graph embedding~\cite{velivckovic2017graph} and machine translation~\cite{vaswani2017attention, bahdanau2014neural}. These models are widely known not only for their improved performance over previous methods but also for their capability to show which input features or learned representations are more important for making a specific prediction. Yang et al. \cite{yang2016hierarchical} use a hierarchical attention network in document classification to capture the informative words as well as the sentences that have a significant role in the decision. This is because the same word or sentence may be differentially important in different contexts. Attention networks also proved to be a useful tool in visual question answering applications, which require a joint image-text understanding to answer a question about the image \cite{xu2015show, lu2017knowing, xu2016ask, lu2016hierarchical}.
Yang et al. \cite{yang2016stacked} propose a Stacked Attention Network (SAN) that uses two attention layers to infer the answer progressively. While the first attention layer focuses on all referred concepts in the question, the higher-level layer provides a sharper attention distribution to highlight regions that are more relevant to the answer.

\textbf{Disentangled Representation Learning.} One goal of representation learning is to break down the features into the independent latent variables that are highly correlated with meaningful patterns~\cite{Goodfellow-et-al-2016}.
In traditional machine learning, approaches such as PCA~\cite{jolliffe2011principal}, ICA~\cite{hyvarinen2000independent} and spectrum analysis~\cite{von2007tutorial} are proposed to discover disentangled components of data.
Recently, deep latent-variable models such as VAE~\cite{kingma2013auto}, InfoGAN~\cite{chen2016infogan} and $\beta$-VAE~\cite{higgins2017beta} were developed to learn disentangled latent variables through variational inference.
For example, in empirical studies, it is shown that $\beta$-VAE and InfoGAN can learn interpretable factorized latent variables of human face images such as azimuth, hairstyle and emotion~\cite{higgins2017beta}.
\subsubsection{Post-hoc Interpretability }\label{subsub:example} Post-hoc interpretable methods aim to explain the decision-making process of the black-box models after they are trained.
These methods map an abstract concept used in a trained machine learning model into a domain that is understandable by humans such as a block of pixels or a sequence of words.
Following are the widely known post-hoc methods. 

\textbf{Local Explanations.} Local Interpretable Model-Agnostic Explanations (LIME) \cite{ribeiro2016should} is a representative and pioneer framework that generates local explanations of black-box models. LIME approximates the prediction of any black-box via local surrogate interpretable models.
LIME selects an instance to explain by perturbing it around its neighborhood (i.e., eliminating patches of pixels or zeroing out the values of some features). These samples are then fed to the complex model for labeling and then it will be weighted based on their proximity to the original data.
Finally, LIME learns an interpretable model on the weighted perturbed data and their associated labels to create the explanations. It is worth noting that LIME is a fast approximation of a broader approach named SHAP \cite{lundberg2017unified} that measures feature importance.

\textbf{Saliency Maps.} Originally introduced by Simonyan et al. \cite{simonyan2013deep} as ``image-specific class saliency maps'', saliency maps highlight pixels of a given input image that are mostly involved in deciding a particular class label for the image. 
To extract those pixels, the derivative of the weight vector is found by a single backpropagation pass (deconvolution).
The magnitude of the derivative shows the importance of each pixel for the class score.
Similar concepts were used by other researchers to deconvolve the prediction and show the locations of the input image that strongly impacts the activation of the neurons \cite{zeiler2014visualizing, springenberg2014striving, selvaraju2017grad}. 
While these methods belong to a popular class of tools for interpretability, Adebayo et al. \cite{adebayo2018sanity} and Ghorbani et al. \cite{ghorbani2019interpretation} suggest that relying on visual assessment is not adequate and can be misleading.

\textbf{Example-Based Explanations.} As proved in education \cite{renkl2014toward} and psychology domains \cite{aamodt1994case}, learning from experiences and examples are promising tools to explain complex concepts.
In these methods, a certain example is selected from the dataset to represent the model's prediction (e.g., k-nearest neighbor) or the distribution of the data. 
It is worth mentioning that example-based explanations should not be confused with those explanations that perturb \emph{features} in the dataset \cite{plumb2018model}.
Although using prototypes as the representation of data has shown to be effective in the human learning process \cite{aamodt1994case}, Kim et al. \cite{kim2016examples} use a method called Maximum Mean Discrepancy (MMD) to capture a more complex distribution of the data.
This method uses some instances as \emph{criticisms} to explain which prototypes are \textit{not} captured by the model to improve the interpretability of the black-boxes. Gurumoorthy et al. \cite{gurumoorthy2017efficient} extend this method and designed a fast prototype selection algorithm called \textit{ProtoDash} to not only select the prototypes and criticism instances, but also output non-negative weights indicating their importance.
% (Kim, Khanna, and Koyejo 2016; Gurumoorthy, Dhurandhar, and Cecchi 2017) use prototypes as example-based explanations

\textbf{Influence Functions.} To track the impact of a training sample on the prediction of a machine learning model, one can simply modify an example or delete it (leave-one-out), retrain the model, and observe the effect. 
However, this approach can be extremely expensive.
To alleviate the issue, influence functions, a classic method from the robust statistics literature, can be used. Koh and Liang \cite{koh2017understanding} proposed a second-order optimization technique to approximate these influence functions.
They verified their technique with different assumptions on the empirical risk ranging from being strictly convex and twice-differentiable to non-convex and non-differentiable losses.\\
Suppose $\hat{y}(x_t,\hat{\theta})$ is the model's prediction for the sample $x_t$ with an optimal parameter $\hat{\theta}$.
Lets $\hat{y}(x_t,\hat{\theta}_{-z})$ be the prediction on the sample $x_t$ when the training sample $z$ was removed while the model's optimal parameter is $\hat{\theta}_{-z}$.
The influence function tries to approximate the difference between the two predictions, $\hat{y}(x_t,\hat{\theta})-\hat{y}(x_t,\hat{\theta}_{-z})$, without retraining the model with the following equation,
\begin{equation}
\label{eq001}
\hat{y}(x_t,\hat{\theta})-\hat{y}(x_t,\hat{\theta}_{-z}) = -\frac{1}{n}\nabla_{\theta}\hat{y}(x_t,\hat{\theta})^TH^{-1}_{\hat{\theta}}\nabla_{\theta}L(z,\hat{\theta}
)
\end{equation}
where $L(z, \hat{\theta})$ is the loss function and $H_{\hat{\theta}}=\frac{1}{n}\nabla^2_{\theta}L(z_i, \hat{\theta})$ is the Hessian matrix.\\
The same authors \cite{koh2019accuracy} also investigate the effect of removing large groups of training points in large datasets on the accuracy of influence functions.
They find out that the approximation computed by the influence functions are correlated with the actual effect.
Inspired by this work, Cheng et al. \cite{cheng2019incorporating} propose an explanation method, \textit{Fast Influence Analysis}, that employs influence functions on Latent Factor Models to resolve the lack of interpretability of the collaborative filtering approaches for recommender systems.

\textbf{Feature Visualization.} 
Another way of describing what the model has learned is feature visualization.
Most methods in this category deal with image inputs. 
Erhan et al. \cite{erhan2009visualizing} present an optimization technique called \emph{activation maximization} to visualize what a neuron computes in an arbitrary layer of  deep neural network. Let $\hat{\theta}$ be the learned fixed parameters after training and $h_{ij}(\hat{\theta}, x)$ be the activation of neuron $i$ in layer $j$, the learned image for that neuron can be calculated by solving the following optimization problem,
\begin{equation}
\label{eq002}
    x^* = \argmax\limits_x h_{ij}(\hat{\theta}, x), \text{ subject to } ||x||_2=1
\end{equation}
Despite this method being used as a tool in providing explanations for higher-layer features \cite{le2013building, olah2018the, Mordvintsev45507}, it has been reported that due to the complexity of the input distribution, some returned images might contain optical illusions \cite{olah2017feature, erhan2009visualizing}.

\textbf{Explaining by Base Interpretable Models.} In section \ref{inr_clssi} we discussed base models such as decision tree, rule-based and linear regression, that are known to be interepretable. Following, we will introduce some works that utilize these algorithms to explain a more sophisticated framework.
Craven and Shavlik \cite{craven1996extracting} are one of the first to use tree-structured representations to approximate neural networks. Since their model is independent of the network architecture and training algorithm, it can be generalized to a wide variety of models. Their method, \textit{TREPAN}, is similar to CART and C4.5 and uses a gain ratio criterion to evaluate the potential splits, but expands the tree based on a node that increases the fidelity of the extracted tree to the network. Inspired by \textit{TREPAN}, Boz \cite{boz2002extracting} propose a method called \textit{DECTEXT} to extract a decision tree that mimics the behavior of a trained Neural Network. In their method, they propose a new splitting technique, a new discretization method, and a novel pruning procedure. With these modifications, the proposed method can handle continuous features, optimize fidelity and minimize the tree size. A technique called distillation \cite{hinton2015distilling} can also be used to fully understand why a specific answer is returned for a particular example. Frosst and Hinton \cite{frosst2017distilling} answer this question by creating a model in the form of soft decision tree and examine all the learned filters from the root of the tree to the classification's leaf node.
Zhang et al. \cite{zhang2019interpreting} adopt the same concept but explained the network knowledge at a human-interpretable semantic level and also showed how much each filter contributes to the prediction.

The MofN algorithm \cite{towell1993extracting} is one of the well-known methods that is used to extracts symbolic rules from trained neural networks. This method clusters the links based on the weights and eliminates those groups that unlikely to have any impact on the consequent. It then forms rules that are the sum of the weighted antecedents with regard to the bias. Authors also report experiments on the fidelity of the model and the comprehensibility of the set rules and the individual rules.

Lou et al. \cite{lou2012intelligible} use a generalized version of linear regression called \textit{generalized additive models} (GAM) in the form of $g(y) = \sum f_i(x_i) = f_1(x_1) + ...+f_n(x_n)$ to interpret the contribution of each predictor for different classifiers or regression models. $g(.)$ is a link function that controls whether we want to describe the model as an additive model (regression by setting $g(y)=y$) or generalized additive model (classification by setting it to a logistic function). $f(.)$ is a shape function that quantifies the impact of each individual feature. This gives the ability to interpret spline models and tree-based shape functions such as single trees, bagged trees, boosted trees and boosted-bagged trees. Due to the model not considering the interactions between the features, there is a significant gap in terms of accuracy between these models and complex models. To fill this gap, the same authors propose a method named \textit{Generalized Additive Models
plus Interactions} (GA$^2$Ms) in the form of $g(y) = \sum f_i(x_i) + \sum f_{ij}(x_i,x_j)$ which takes into account the two-dimensional interactions that still can be interpretable as heat maps \cite{Lou:2013:AIM:2487575.2487579}. Two case studies are conducted on real healthcare problems on predicting pneumonia risks by using GA$^2$Ms. These studies uncover new patterns that are ignored by state-of-the-art complex models while still hitting their accuracy \cite{Caruana:2015:IMH:2783258.2788613}.

% \subsubsection{Traditional Model based interpretability}

% %Ruocheng : 
% %Explanation-producing networks.  -> 
% %Attention Networks
% %Disentangled Representations
% %Generate Explanations

% \noindent\textbf{Explanation-producing Networks.}
% To improve the interpretability of the black-box deep neural networks, a variety of methods are proposed.
% %
% Following~\cite{gilpin2018explaining}, we provide a brief introduction to three classes of approaches which produce explanations for deep learning models: (1) attention networks, (2) learning disentangled representations and (3) generating explanations.
% %

% %

%Raha:
%1) Provide justification of outputs: emulate processing of data to connect inputs and outputs: 
%linear proxy models; 
%decision trees; 
%automatic-rule extraction; 
%Salience mapping

%2) Explain representations of data in NN.
%Layers
%Individual Units
%Representation Vectors

% \begin{itemize}
%     \item Interpretable models
%     \item transparency of the model 
%     \item Post-hoc (model agnostic) interpretability
% \end{itemize}
% \subsubsection{Traditional example based interpretability}
%Roughly 1.5 pages

% \begin{itemize}
%     \item Prototype Samples
%     \item Influential instances 
%     \item Adversarial examples
% \end{itemize}

%one paragraph each and their problems and motivation behind causal interpretation
\section{Causal Inference} \label{causal}
%\raha{connection between traditional and causal}
In this section, we briefly review the concepts from causal inference used in this paper for causal interpretable models. In their paper, Guo et al. \cite{guo2018survey} provide a comprehensive review of existing causal inference methods and definitions.
%causal inference concepts and definitions go here!

\begin{definition}[Structural Causal Models ]

 A 4-tuple variable $M(X, U, f, P_u)$ where X is a finite set of endogenous variables, usually the observable variables, U denotes a finite set of exogenous variables which usually account for unobserved or noise variables, f is a set of function $\{f_1, f_2, ..., f_n\}$ where each function represents a causal mechanism such that $\forall x_i \in X, x_i = f_i(Pa(x_i), u_i)$ and $Pa(x_i)$ is a subset of ~~($X\setminus\{x_i\})  \cup  U$ and $P_u$ is a probability distribution over U  is called An Structural Causal Model (SCM) or Structural Equation Model (SEM)\cite{pearl2009causality}. 
\end{definition}
\begin{definition} [Causal Bayesian Network]

To represent an SCM $M(X, U, f, P_u)$, a directed graphical model $G(V, E)$ is used. V is the set of endogenous variables X and E denotes the causal mechanisms. This indicates for each causal mechanism $x_i = f_i(Pa(x_i), u_i)$, there exists a directed edge from each node in the parent set $Pa(x_i)$ to $x_i$. The entire graph representing this SCM is called a Causal Bayesian Network (CBN).
\end{definition} 

\begin{definition}[Average Causal Effect] 

The Average Causal Effect (ACE) of a binary random variable x (treatment) on another random variable (outcome) is defined as:
\begin{equation}
\label{eq:equation1}
\begin{aligned}
ACE = \EX[y|do(x = 1)] - \EX[y|do(x = 0)],
\end{aligned}
\end{equation}

Where do(.) operator denotes the corresponding interventional distribution defined by the SCM or CBN.
\end{definition}

\section{Causal Interpretablity} \label{causalinterpret}
In this section, we discuss the state-of-the-art frameworks on causal interpretability. These frameworks are particularly needed since objective functions of machine learning models only capture correlations and not real causes. Therefore, these models might cause problems in real-world decision making, such as making policies related to smoking and cancer. Moreover, training data used to train these models might not perfectly represent the environment; and the train and the test sets might also have different distributions. A causal interpretable model can help us understand the real causes of decisions made by machine learning algorithms, improve their performance, and prevent them from failing in unexpected circumstances.

%The motivations behind these models are as follows:
% Motivations:
%%%%%%%%%%%%%%%%%%%%%% ATTENTION RAHAAAAAAAAAAAA %%%%%%%%%%%%%%%%%
% ;)
%%%%%%%%%%%% BULLETS %%%%%%%%%%
% :p
%\begin{itemize}
%    \item The objective function of a machine learning model may only capture correlations but not real causes. Such models may not be useful for real-world decision making such as making policies related to smoking and cancer.
%    \item Training data may imperfectly represent the environment. This can lead to different distributions between the training and test sets. 
    % (domain adaptation).
%\end{itemize}
%\raha{Difference between traditional and causal approaches}
Pearl \cite{DBLP:journals/corr/abs-1801-04016} introduces different levels of said interpretability and argues that generating counterfactual explanations is the way to achieve the highest level of interpretability. Below are those levels of interpretability and their definitions:
%\subsubsection{Desiderata of a causal interpretable model}
%\cite{DBLP:journals/corr/Lipton16a} \cite{DBLP:journals/corr/abs-1801-04016}
\begin{itemize}
    \item Statistical (associational) interpretability: Aims to uncover statistical associations by asking questions such as ``How would seeing $x$ change my belief in $y$?''
    \item Causal interventional interpretability: Is designed to answer ``What if'' questions. 
    \item Counterfactual interpretability: Is the highest level of interpretability, which aims to answer ``Why'' questions.
    
\end{itemize}

Traditional interpretability mainly focuses on the statistical interpretability, whereas causal interpretability aims to answer questions associated with the causal interventional interpretability and counterfactual interpretability.
In the following, we provide an extensive review of existing work on causal interpretability. We classify the existing works in this field into four main categories:
\begin{enumerate}
    \item Causal interpretablity for model-based interpretations: In this category, methods explain the causal effect of a model component on the final decision.
    \item Counterfactual explanation generators: Methods in this category aim to generate counterfactual explanations for alternate situations and scenarios.
    
    \item Causal interpretability and fairness: Lipton \cite{DBLP:journals/corr/Lipton16a} explains that interpretable models are often indispensable to guarantee fairness. Motivated by this, we provide an overview of the state-of-the-art methods on causal fairness.
    \item Causal interpretability and its role in verifying the causal relationships discovered from data: In this category, we review methods which leverage interpretability as a tool to verify causal assumptions and relationships. We also discuss the scenarios, where causal inference can be used to guarantee the interpretability of a machine learning model.

\end{enumerate}
In the following, we discuss each category in detail.
\subsection{Causal Inference and Model-based Interpretation}
Recently, causality has gained increasing attention in explaining machine learning models \cite{DBLP:journals/corr/abs-1902-02302, DBLP:journals/corr/abs-1802-00541}. 
These approaches are usually designed to explain the role and importance of each component of a machine learning model on its decisions with concepts from the causality.
% For instance, estimating the causal effect of a neuron on the final decision of a neural network model by estimating the ACE of the neuron on the output.
For instance, one way to explain the role of a neuron on the decision of a neural network is to estimate the ACE of the neuron on the output \cite{DBLP:journals/corr/abs-1902-02302, parafita2019explaining}. Traditional interpretable models cannot answer vital questions for understanding machine learning models. For instance, traditional machine interpretability frameworks are not capable to answer causal questions such as ``What is the impact of the n-th filter of the m-th layer of a deep neural network on the predictions of the model?'' which are helpful and required for understanding a neural network model.
Furthermore, despite being simple and intuitive, performing ablation testing (i.e., removing a component of the model and retraining it to measure the performance for a fixed dataset) is computationally expensive and impractical. 
%
%We present two main motivations of creating causal interpretations for machine learning models:
%\begin{enumerate}
%    \item Causal questions which are useful and necessary for understanding a machine learning model have not been answered by traditional machine interpretability frameworks. For instance, traditional machine interpretability frameworks are not capable to answer causal questions such as \emph{what is the impact of the n-th filter of the m-th layer of a deep neural network on the predictions of the model?} which are useful and necessary for understanding a neural network model.
%    \item Despite being simple and intuitive, performing ablation testing (i.e., removing a component of the model and retraining it to measure the performance for a fixed dataset) is computationally expensive and impractical.
    %\item Traditional attribution methods do not satisfy the axioms of evaluation of methods.
%\end{enumerate}
To address these problems, causal interpretability frameworks have been proposed. %\rg{maybe cite some reference here?}. \raha{I think we already refrenced them too many times}
These frameworks are mainly designed to explain the importance of each component of a deep neural network on its predictions by answering counterfactual questions such as ``What would have happened to the output of the model had we had a different component in the model?''. 
These types of questions are answered by borrowing some concepts from the causal inference literature. 
The main idea is to model the structure of the DNN as a SCM and estimate the causal effect of each component of the model on the output by performing causal reasoning. 
Narendra et al. \cite{DBLP:journals/corr/abs-1811-04376} consider the DNN as an SCM, apply a function % \rg{what are desired function and value?} 
on each filter of the model to obtain the targeted value such as variance or expected value of each filter and reason on the obtained SCM. Harradon et al. \cite{DBLP:journals/corr/abs-1802-00541} further suggest that in order to have an effective interpretability, having a human-understandable causal model of DNN, which allows different kinds of causal interventions, is necessary.
Based on this hypothesis, the authors propose an interpretability framework, which extracts human-understandable concepts such as eyes and ears of a cat from deep neural networks, learns the causal structure between the input, output and these concepts in an SCM and performs causal reasoning on it to gain more insights into the model.
Chattopadhyay et al. \cite{DBLP:journals/corr/abs-1902-02302} propose an attribution method based on the first principle of causality, particularly SCMs and $do(\cdot)$ calculus. 
More concretely, similar to other proposed methods in this category, the proposed framework models the structure of the machine learning algorithm as an SCM. It then proposes a scalable causal inference approach to the estimate individual treatment effect of a desired component on the decision made by the algorithm.

Chattopadhyay et al. suggest to simplify the SCM defined on a multi-layer network $M([l_1, l_2, l_3 ...., l_n],U, f, P_U)$ to another network as SCM $M'([l_1, l_n],U, f', P_U)$ where $l_1$ and $l_n$ represent neurons in the input and output layers, $l_i$ represents neurons in the i-th layer of the network, $U$ denotes the set of unknown variables, $f$ and $f'$ correspond to the SCM functions and $P_U$ defines distributions of the unknown variables. %\rg{you may also need to cover the meaning of $U$, $f,f'$ and $P_U$}. 
They then propose to calculate the ACE %\rg{did we define average causal effect before?} 
of any neurons of the model on the output by performing causal reasoning on \textit{M} as follows,
\begin{equation}
\label{eq:equation1}
\begin{aligned}
ACE_{do(x_i = \alpha)}^{y} = \EX[y|do(x_i = \alpha)] - baseline_{x_i},
\end{aligned}
\end{equation}
where $x_i$ is $i$-th neuron of the network, $y$ is the output of the model and $\alpha$ is an arbitrary value the neuron is set to. They also propose to calculate the $baseline_{x_i}$ as $\EX_{x_i}[\EX_{y}[y|do(x_i = \alpha)]]$ %\rg{please explain $\alpha$, $x_i$ and $y$}.
%Another line of research leverages causal inference to explain the model in the case that model structure can not be considered a SEM. \cite{alvarez-melis-jaakkola-2017-causal} and \cite{DBLP:journals/corr/abs-1905-10958} are examples of this category of models. \cite{alvarez-melis-jaakkola-2017-causal} proposes a causal framework to explain predictions of black-box sequence-to-sequence models by 

%Second category represents the models which are designed to perform causal inference tasks\raha{briefly mention what they are} and are inherently interpretable.

In another research direction, Zhao and Hastie \cite{zhao2019causal} state that to extract the causal interpretations from black-box models, one needs a model with good predictive performance, domain knowledge in the form of a causal graph, and an appropriate visualization tool. They further explore partial dependence plot (PDP) \cite{friedman2001greedy} and Individual Conditional Expectation (ICE) \cite{goldstein2015peeking} to extract causal interpretations from black-box models.
Alvarez-Melis and Jaakkola \cite{alvarez-melis-jaakkola-2017-causal} generated causal explanations for structured input structured output black-box models by (a) generating perturbed samples using a variational auroencoder; (b) generating a weighted bipartite graph $G = (V_x \cup V_y, E)$, where $V_x$ and $V_y$ are elements in x and y and $E_{ij}$ represents the causal influence of $x_i$ and $y_j$; and (c) generating explanation components using graph partitioning algorithms.

Parafita and Vitria \cite{parafita2019explaining} introduce a causal attribution framework to explain decisions of a classifier based on the latent factors. The framework consists of three steps, (a) constructing Distributional Causal Graph which allows us to sample and compute likelihoods of the samples; (b) generating a counterfactual image which is as similar as possible to the original image; and (c) estimating the effect of the modified factor by estimating the causal effect.

Causal interpretation has also gained a lot of attention in Generative Adversarial Networks (GANs) interpretability. Bau et al. \cite{DBLP:journals/corr/abs-1811-10597} propose a causal framework to understand "How" and "Why" images are generated by Deep Convolutional GANs (DCGANs). This is achieved by a two-step framework which finds units, objects or scenes that cause specific classes in the data samples. In the first step, dissection is performed, where classes with explicit representations in the units are obtained by measuring the spatial agreement between individual units of the region we are examining and classes using a dictionary of object classes. In the second step, intervention is performed to estimate the causal effect of a set of units on the class.
This framework is then used to find the units with the highest causal effect on the class. Following equation shows the objective of this framework,
\begin{equation}
\label{eq16}
\alpha^* = \argmin_{\alpha}(-\delta_{\alpha\rightarrow c} + \lambda||\alpha||_2),
\end{equation}
where $\alpha$ indicates the units that have causal effect on the outcome, $\delta_{\alpha\rightarrow c}$ measures the causal effect of units on the class by intervening on $\alpha$ and set it to the constant c and $\lambda||\alpha||_2$ is a regularization term.
Besserve et al. \cite{DBLP:journals/corr/abs-1812-03253} propose to better understand the internal functionality of generative models such as GANs or Variational Autoencoders (VAE) and answer questions like "For a face generator, is there an internal encoding of the eyes, independent of the remaining facial features?", by manipulating the internal variables using counterfactual inference.

Madumal et al. \cite{DBLP:journals/corr/abs-1905-10958} leverage causal inference to explain the behavior of reinforcement learning agents by learning an SCM during reinforcement learning and generate counterfactual examples using the learned SCM.
\subsection{Causal Inference and Example-based Interpretation}
As mentioned in Section \ref{traditional}, in example based explanations, we are looking for data instances that are capable of explaining the model or the underlying distribution of the data. 
In this subsection, we explain \emph{counterfactual explanations}, a type of example-based explanations, which are one of the widely used explanations for interpreting a model's decisions. Counterfactual explanations aim to answer ``Why'' questions such as ``Why the model's decision is Y?'' or ``Was it input X that caused the model to predict Y?''. % {\color{blue} add some reference for counterfactuals} 
Generally speaking, counterfactuals are designed  to answer hypothetical questions such as ``What would have happened to Y, had I not done X?''.
They are designed based on a new type of conditional probability $P(y_x|x', y')$. 
%
%This probability indicates how the probability of real life observation, i.e.  $x', y'$ would change to $y_x$ if $x'$ is set to $x$. 
This probability indicates how likely the outcome (label) of an observed instance, i.e., $y'$, would change to $y_x$ if $x'$ is set to $x$.
These kinds of questions can be answered using SCMs \cite{article}.

%\raha{add some explanations about counterfactuals from causal perspective}.
%
%{\color{blue}add some reference for counterfactual examples} 
Counterfactual explanations are defined as examples that are obtained by performing minimal changes in the original instance's features and have a predefined output.
For example, what minimal changes can be made in a credit card applicant's features such that their application gets accepted.
 These explanations are human friendly because they are usually focused on a few number of features and therefore are more understandable. However, they suffer from the \textit{Roshomon} effect \cite{molnar2019} which means there could be multiple true versions of explanations for a predefined outcome. 
To alleviate this problem, we could report all possible explanations, or find a way to evaluate all explanations and report the best one. 
%A good explanation can be defined by the following two criteria: (1) The outcome of the counterfactual explanation should be as close as possible to the predefined output (the alternative reality) (2) Features of the counterfactual example should be as close as possible to those of the original instance. 
%\raha{Example?} One example of application of counterfactual explanations is explanations of loan applications. 
%
Recently, several works have been proposed to generate counterfactual explanations.
In order to generate counterfacutal examples, Wachter et al. \cite{DBLP:journals/corr/abs-1711-00399} propose to minimize the mean squared error between the model's predictions and counterfactual outcomes as well as the distance between the original instances and their corresponding counterfactuals in the feature space.
Eq. \eqref{eq:equation1} shows the objective function to achieve this goal,
\begin{equation}
\label{eq:equation1}
\begin{split}
&\argmin_{x_{cf}} \max_{\lambda}L(x, x_{cf}, y, y_{cf})\\
& L(x, x_{cf}, y, y_{cf}) = \lambda \cdot (\hat{f}(x_{cf}) - y_{cf})^2 + d(x, x_{cf}),
\end{split}
\end{equation}
where the first term indicates the distance between the model's prediction for the counterfactual input $x_{cf}$ and the desired counterfactual output, while the second term indicates the distance between the actual instance features x and the counterfactual features $x_{cf}$.
Liu et al. \cite{DBLP:journals/corr/abs-1907-03077} propose a generative model to generate counterfactual explanations for explaining a model's decisions using Eq.\eqref{eq:equation1}.
 Garth et al. \cite{DBLP:journals/corr/abs-1811-05245} propose a method to generate counterfactual examples in a high dimensional setting. The method is proposed for credit application prediction via off-the-shelf interchangeable black-box classifiers. In the case of high dimensional feature space, the generated explanation might not be interpretable due to the existence of too many features. 
To alleviate the problem, the authors propose to reweigh the distance between the features of an instance and its corresponding counterfactual with the inverse median absolute deviation (Eq.\eqref{eq:equation2}). This metric is robust to outliers and results in more sparse, and therefore, more explainable solutions.
\begin{equation}
\label{eq:equation2}
MAD_j = median _{i \in \{1, 2, ..., n\}} (|x_{i,j} - median_{l \in \{1, 2, ..., n\}}(x_{l,j})|)
\end{equation}

Goyal et al. \cite{DBLP:journals/corr/abs-1904-07451} propose to generate counterfactual visual explanations for a query image $I$ by using a distractor image $I'$ which belongs to the class $c'$ (a different class from the actual output of the classifier). 
To generate counterfactual explanations, the authors propose to detect spatial regions in $I$ and $I'$ such that replacing those regions in $I$ with regions in $I'$ results in system classifying the generated image as $c'$. 
In order to avoid trivial solutions such as replacing the entire image $I$ with $I'$, authors propose to minimize the number of edits to transform $I$ to $I'$. 
The proposed framework is shown in the following equation,
\begin{equation}
\label{eq:equation3}
\begin{split}
&\underset{P, a}{\text{min}}~ ||a||_1\\
& \text{s.t.}~ c' = \text{argmax}~g((\mathbf{1}-a)\circ f(I) + a\circ Pf(I'))\\
&a_i \in \{0, 1\}~ \forall i ~\text{and}~ P \in \mathcal{P},\\
\end{split}
\end{equation}
where $a \in \mathbb{R}^{hw}$ ($h$ and $w$ represent height and width of an image, respectively) is a binary vector which indicates whether the feature in $I$ needs to be changed with the feature in $I'$ (value 1) or not (value 0). 
$ P \in \mathbb{R}^{hw\times hw}$ is a permutation matrix used to align spatial cells of $f(I')$ with $f(I)$, $f(I)$ and $f(I')$ correspond to spatial feature maps of $I$ and $I'$, respectively. Function $g(.)$ represents the classifier and $\mathcal{P}$ is a set of all $hw\times hw$ permutation matrices.
Goyal et al. \cite{DBLP:journals/corr/abs-1907-07165} propose to explain classifiers' decisions by measuring the \emph{Causal Concept Effect} (CACE). CACE is defined as the causal effect of a concept (such as the brightness or an object in the image) on the prediction. In order to generate counterfactuals, authors leverage a VAE-based architecture.
Hendricks et al. \cite{DBLP:journals/corr/abs-1812-01263} propose a method to generate counterfactual explanations using multimodal information for video classification tasks. 
The proposed method in this work generates visual-linguistic explanations in two steps. First, it trains a classification model for which we would like to generate explanations. Then, in the second step, it trains a post-hoc explanation model by leveraging the output and mid-level features of the trained model in first the step. The explanation model predicts the counterfactuality score for all the negative classes (classes that the instance does not belong to according to the prediction model trained in the first step). 
The explanation model then generates explanations by maximizing the counterfactuality score between positive and negative classes.
%

%\cite{DBLP:journals/corr/abs-1806-09809} proposes a method to generate counterfactual explanations using natural language.
%
%The paper proposes to examine features that are missing but could have contributed to the output.
%\rg{I could not understand the above sentence. Maybe you can provide more details.}
%
Moore et al. \cite{DBLP:journals/corr/abs-1906-10671} propose to leverage adversarial examples to generate counterfactual explanations.
In order to generate plausible explanations, the number of changed features should be small. Moreover, some features such as age cannot be changed arbitrarily. For example, we cannot ask loan applicants to reduce their age.  Therefore, to constrain the number of changed features and the direction of gradients in the generated adversarial examples, authors propose to mask the unwanted features and gradients in a way that only desired features change in the generated explanations.

Kommiya et al. \cite{DBLP:journals/corr/abs-1905-07697} propose to explain the decision of a machine learning framework by generating counterfactual examples which satisfy the following two criteria, (1) generated examples must be feasible given users conditions and context such as range for the features or features to be changed; (2) counterfactual examples generated for explanations should be as diverse as possible. 
In order to impose the diversity criterion, authors propose to either maximize the point-wise distance between examples in feature-space or leverage the concept from Determinantal point processes to select a subset of samples with the diversity constraint. 

Van Looveren and Klaise \cite{DBLP:journals/corr/abs-1907-02584} propose to leverage class prototypes to generate counterfactual explanations. 
They also claim that using class prototypes for counterfactual example generation accelerates the process. This work suggests that the generated examples by traditional counterfactual generation frameworks \cite{DBLP:journals/corr/abs-1711-00399, DBLP:journals/corr/abs-1811-05245} do not satisfy two main criteria: (1) they do not consider the training data manifold which may result in out-of-distribution examples, and (2) the hyperparameters in the framework should be carefully tuned in an appropriate range which could be time consuming. To solve the mentioned problems, the authors propose to add a reconstruction loss term (defined as $L_2$ reocnstruction error between counterfactuals and an autoencoder trained on the training samples) as well as a prototype loss term, which is defined as $L_2$ loss between the class prototype and the counterfactual samples, to the original objective function of counterfactual generation (Eq \eqref{eq:equation1}).

%\begin{equation}
%\label{eq14}
%L_{AE} = \gamma||x_0 + \delta − AE(x_0 + \delta)||_2^2
%\end{equation}
%Where AE is an autoencoder trained on the training data and $L_{AE}$ represents the reconstruction error loss for the counterfactual instance.

Rathi \cite{DBLP:journals/corr/abs-1906-09293} generates counterfactual explanations using shapely additive explanations (SHAP).

Hendricks et al. \cite{DBLP:journals/corr/abs-1806-09809} defined a method to generate natural language counterfactual explanations. The framework checks for evidences of a counterfactual class in the text explanation generated for the original input. It then checks if those factors exist in the counterfactual image and returns the existing ones. 
%\raha{TODO: compare counterfactual examples with other types of example based methods such as prototype}

%\subsubsection{\textbf{Interpretability and verification of causal inference}}

\subsection{Causal Inference and Fairness} \label{fairness}

Nowadays, politicians, journalists and researchers are concerned regarding the interpretability of model's decisions and whether they comply with ethical standards \cite{goodman2016regulations}.
Algorithmic decision making has been widely utilized to perform different tasks such as approving credit lines, filtering job applicants and predicting the risk of recidivism \cite{DBLP:journals/corr/Chouldechova17}. %i.e., the probability of an individual relapses into criminal behavior \cite{DBLP:journals/corr/Chouldechova17}.
Prediction of recidivism is used to determine whether to detain or free a person and therefore, it needs to be guaranteed that it does not discriminate against a group of people. 
Since conventional evaluation metrics such as accuracy does not take these into account, it is usually required to come up with interpretable models in order to satisfy fairness criteria.
Recently, huge attention has been paid to incorporating fairness into decision making methods and its connection with causal inference.
 Kusner et al. \cite{DBLP:conf/nips/KusnerLRS17} propose a new metric for measuring how fair decisions are based on counterfactuals. 
According to this paper, a decision is fair for an individual if the outcome is the same both in the actual world and a counterfactual world in which the individual belonged to a different demographic group.
Kilbertus et al. \cite{NIPS2017_6668} address the problem from a data generation perspective by going beyond observational data. The authors propose to utilize causal reasoning to address the fairness problem by asking the question ``What do we need to assume about the causal data generating process?'' instead of ``What should be the fairness criterion?''.
Madras et al. \cite{DBLP:journals/corr/abs-1809-02519} propose a causal inference model in which the sensitive attribute confounds both the treatment and the outcome. It then leverages deep learning techniques to learn the parameters of the model.
Zhang and Bareinboim \cite{inproceedings} propose a metric (i.e., causal explanations) to quantitatively measure the fairness of an algorithm. This measure is based on three measures of transmission from cause to effect namely counterfactual direct (Ctf-DE), indirect (Ctf-IE), and spurious (Ctf-SE) effects as defined below.
Given an SCM  $M$, the counterfactual indirect effect of intervention $X = x_1$ on $Y = y$ (relative to baseline $X = x_0$) conditioned on $X = x$ with mediator $W = W_{x_1}$ is defined as,
\begin{equation}
\label{eq:equation4}
IE_{x_0, x_1} (y|x) = P(y_{x_0,W_{x_1}} |x) - P(y_{x_0} |x)
\end{equation}

the counterfactual direct effect of intervention $X = x_1$ on Y (with baseline $x_0$) conditioned on $X = x$ is defined as,
\begin{equation}
\label{eq:equation5}
DE_{x_0, x_1} (y|x) = P(y_{x_1,W_{x0}} |x) - P(y_{x_0} |x)
\end{equation}

And finally, the spurious effect of event $X = x_1$ on $Y = y$ (relative to baseline $x_0$) is defined as,
\begin{equation}
\label{eq:equation6}
SE_{x_0, x_1} (y|x) = P(y_{x_0} |x_1) - P(y |x_0)
\end{equation}

\subsection{Causal Inference as Guarantee for Interpretability}
%\raha{it could belong to transparency of the model?}
Machine learning has had great achievements in medical, legal and economic decision making. Frameworks for these applications must satisfy the following two criteria: 1) they must be causal 2) they must be interpretable. For example, in order to find the efficacy of a drug on patient's health, one needs to estimate the causal effect of the drug on patient's health status. Moreover, in order for the results to be reliable for doctors and experts, an explanation of how the decision has been made is necessary. 
%\rg{you may want to explain these two criteria matter using examples (maybe a running example throughout the paper).}
%
Despite recent achievements in these two fields separately, not so many works have been done to cover both requirements simultaneously.
Moreover, the state-of-the-art approaches in each field are incompatible and therefore can not be combined and used together.
Kim and Bastani~\cite{DBLP:journals/corr/abs-1901-08576} propose a framework to bridge the gap between causal and interpretable models by transforming \textit{any} algorithm into an interpretable individual treatment effect estimation framework. 
To be more specific, this work leverages the algorithm proposed in \cite{shalit2017estimating} to learn an oracle function $f$ which estimates the causal effect of a treatment for any observed instance and then learn an interpretable function $f'$ to estimate $f$. They further provide a bound for the error produced by their framework.
 
In another line of research, causal interpretability has been used to verify the causal relationships in the data.
Caruana et al. \cite{Caruana:2015:IMH:2783258.2788613} perform two case studies to discover the rules which show cases where generalized additive models with pairwise interactions ($GA^2Ms$) learn rules based on only correlations in the data and invade causal rules. They then propose to fix the learned rules based on domain experts knowledge.
%Recently, causal interpretability has be

Bastani et al. \cite{ribeiro2016model} propose a decision tree based explanation method to generate global explanations for a black-box model. Their proposed framework provides powerful insights into the data such as causal issues confirmed by the physicians previously.
%\raha{causal guarantee paper can be used as a bridge between two fields}
\section{Performance evaluation} \label{eval}
%\raha{Template: (i) what is this section for, (ii) why it is important, (iii) what key elements are, and (iv) how the evaluation is done, and (v) what are challenges including how it is different from the conventional ML evaluation}

In this section we provide a detailed review of evaluation methods and common datasets used to assess the interpretability of models for causal interpretablity. Evaluation of interpretability is a challenging task due to the lack of consensus definition of interpretability and understanding of humans from the concept. Evaluation of causal interpretability is even more challenging due to the lack of groundtruth data for causal explanations and verification of causal relationships. Therefore, it is important to have a unified guideline on how to evaluate the proposed models. 
Traditional interpretability of a model is usually measured with quantifiable proxies such as if a model is approximated using sparse linear models it can be considered interpretable. To evaluate the causal interpretability, researchers also came up with some proxy metrics such as size and diversity of the counterfactual explanation. In this section, we discuss all criteria defined for the ``goodness'' of both causal and traditional interpretations and proxy metrics to measure how good the proposed framework can generate these explanations.
\subsection{Datasets}

%\cite{DBLP:journals/corr/abs-1812-01263}
%\subsubsection{Traditional interpretability datasets}
In this section, we briefly introduce benchmark datasets commonly used to evaluate interpretable models. Depending on the the type of the data (i.e., text, image or tabular) different datasets are used to assess the interpretability . Some commonly used datasets for image are ``ImageNet (ILSVRC)'' \cite{ILSVRC15}, ``MNIST'' \cite{mnist_handwritten_digit_database} and ``PASCAL VOC dataset'' \cite{Everingham10}. While for text they experimented on ``20 Newsgroup Dataset'' \cite{20_NEWS}, ``Yelp'' \cite{yelp_dataset}, ``IMDB'' \cite{imdb} and ``Amazon'' \cite{amazon} reviews. ``UCI repository'' \cite{uci_repository} consists of some tabular datasets that were used by the litreture such as ``Spambase'', ``Insurance'', ``Magic'', ``Letter'', and ``Adult'' datasets.
In order to explain the outcome of the test sample, the explanations are provided by the model. For instance, in the case of image data, those patches of the images that are mostly responsible for the class label were selected. For the text data, words involved in the final decision are made bold with different shades of color, which represent the degree of their involvement. 
In addition to the mentioned datasets, there are some datasets commonly used to evaluate the causal interpretable frameworks. In the following, we list common datasets used for the evaluation of causal interpretability.
\begin{itemize}
%\item CACE datasets
\item \textit{German loan dataset} \cite{Dua:2019}. This dataset contains 1000 observations of loan applicants which contains, numeric, categorical and ordinal attributes.

\item \textit{LendingClub}.
This dataset\footnote{\url{https://www.lendingclub.com/info/download-data.action}} contains 5 years of loan records (2007-2011) given by LendingClub company. After preprocessing, it contains 8 features, namely, employment years, annual income, number of open credit accounts, credit history, loan grade
as decided by LendingClub, home ownership, purpose, and the state
of residence in the United States.
. 
\item \textit{COMPAS}. Collected by ProPublica \cite{article12} for analysis purposes on recidivism decisions in the United States, after preprocessing, this data contains 5 features, namely, bail applicant's age, gender, race, prior count
of offenses, and degree of criminal charge. 
%\item \textbf{HELOC (Home Equity Line of Credit)} Used in the FICO 2018 xML Challenge \cite{}, this data contains real anonymized credit applications by homeowners. 

\end{itemize}
 Unfortunately, datasets used for this purpose are not specifically designed for causal interpretability and do not contain the groundtruth that captures the causal aspect of the model such as counterfactual explanations or the ACE of different components of the model on the final decision. 
On the other hand, there are existing benchmark datasets specifically designed for evaluating tasks in causal inference. Cheng et al. \cite{li2019causal} provide a comprehensive survey on benchmark datasets for different causal tasks.

\begin{table*}[]
\begin{tabular}{|c|m{8em}|m{15em}|m{25em}|}
\hline
\multicolumn{1}{|l|}{} & \textbf{Countrfactual Property } & \textbf{Description of Property} & \textbf{Evaluation Metrics} \\ \hline \hline
\multirow{2}{*}{1}   & \multirow{2}{*}{Sparsity/Size}  & \multirow{2}{15em}{Perturbation which transforms $x$ to $x_{cf}$ should be small}  & Elastic net loss term ($EN(\delta) = \beta. ||\delta||_1 + ||\delta||_2^2$) \cite{DBLP:journals/corr/abs-1907-02584}\\ \cline{4-4} &  &  & Counting number of altered features manually \cite{DBLP:journals/corr/abs-1811-05245}\\ \hline
\multirow{2}{*}{2}  & \multirow{2}{*}{Interpretability}  & \multirow{2}{15em}{Counterfactual explanations should lie close to data manifold} & Ratio of the reconstruction errors of counterfactual generator trained only on the counterfactual class and counterfactual generator trained on the \emph{original} class \cite{DBLP:journals/corr/abs-1907-02584} \\ \cline{4-4} & & & Ratio of the reconstruction errors of counterfactual generator trained only on the counterfactual class and counterfactual generator trained on the \emph{all} class \cite{DBLP:journals/corr/abs-1907-02584}\\ \hline 3 & Proximity & Counterfactual explanations should be as similar as possible to the original instance  & $Proximity= -\frac{1}{k}\sum_{i=1}^{k}dist(x_{cf_i}, x)$ \cite{DBLP:journals/corr/abs-1905-07697} \\ \hline 4  & Speed  & Generating counterfactuals should be fast enough to be deployable in real-world applications & Measure the time and number of gradient updates \cite{DBLP:journals/corr/abs-1907-02584} \\ \hline 5 & Diversity  & Counterfatual explanations generated for a data instance should be different from each other & $Diversity = \frac{1}{|C_k|^2} \sum_{i=1}^{k-1} \sum_{j=i+1}^k dist(x_{cf_i}, x_{cf_j})$ \cite{DBLP:journals/corr/abs-1905-07697} \\ \hline
\multirow{2}{*}{6}     & \multirow{2}{*}{\begin{tabular}[c]{@{}l@{}}Visual-Linguistic \\ Counterfactuals\end{tabular}} & Visual explanation is the region which retains high positiveness or negativeness (i.e.,  on the model prediction for specific positive or negative classes). & Measure how the output of the target classifier changes corresponding to the negative class when a specific region is removed from the input using accuracy \cite{DBLP:journals/corr/abs-1812-01263}. \\ \cline{3-4} &  & Linguistic explanation is compatible to the visual counterpart.  & Measure how the output of the target classifier changes corresponding to the negative class when a specific region is removed from the input using accuracy \cite{DBLP:journals/corr/abs-1812-01263}.  \\ \hline
\end{tabular}
\caption{A summary of evaluation metrics for counterfactual explanations}
\label{table:counteval}
\end{table*}

\subsection{Evaluation Metrics}
%Evaluation of Interpretable models is a challenging tasks for both traditional and causal models. However, this is even more challenging for causal approaches since verifying the causality of the proposed model is itself challenging.

In order to assess the performance of a causal interpretable framework, authors are required to evaluate the interpretability of generated explanations from two aspects, (1) the quality of the generated explanations, i.e., are generated explanations interpretable to humans?; and (2) are the generated explanations causal? 
In the following two subsections, we provide comprehensive guidelines and metrics on how to answer these questions.

\subsubsection{Interpretability Evaluation Metrics}
Evaluating the interpretability of a machine learning model is usually a challenging task.
Interpretable frameworks often evaluate their methods via two main perspectives, (1) how well the generated explanations by the method match the human expectation from different aspects; (2) how well the generated explanations are without using any human subjects. Thus, we will categorize different assessment methods based on the aforementioned perspectives and provide some examples of experiments conducted by the researchers.

\textbf{Human Subject-Based Evaluation Metrics.}
Part of the research in interpretability aims to let humans understand the reasons behind the outcome of a product. Accordingly, experiments carried out by the researchers usually answer the following questions:
\begin{itemize}
    \item By providing two different models, can the explanations help users choose the better classifier in terms of generalizability? This will help us to investigate whether the explanations can be used to decide which model is better. Ribeiro et al. \cite{ribeiro2016should} used human subjects from ``Amazon Mechanical Turk'' (AMT) to choose between two models, one that generalizes better than the other while its accuracy was lower on cross validation. With the provided explanations, the subjects were able to choose the more generalized model 89\% of the time.
    \item With explanations provided by the interpretable methods for a particular sample, can a user correctly predict the outcome of that sample? This is also called ``Forward Simulation/Prediction'' by  Doshi-Velez and Kim \cite{doshi2017towards}. We can verify the explanations actually defines the output we are looking for.
    \item Based on the explanations, do users trust the classifier to be used in real-world applications? Selvaraju et al. \cite{selvaraju2017grad} evaluated the trust by asking 54 AMT workers to rate the reliability of the models via a 5-point scale questionnaire. A sample along with its explanations were demonstrated to subjects for two different models, AlexNet and VGG-16 (VGG-16 is known to be more reliable than AlexNet). Moreover, only those instances that provided the same prediction and were aligned with the ground truth label were considered. The results of the evaluation shows that with the proposed explanation the subjects trust the model that generalizes better (VGG-16).
    \item Do the resulted explanations match human intuition? The model is described to human subjects in detail and they were asked to provide insights about the outcome of the model (human-produced explanations). The test assumes that the explanations provided by the human should be aligned with one that the model provides \cite{lundberg2017unified}. Moreover, experts in a specific field (e.g., doctors) can also be used to provide the explanations (e.g., important factors/symptoms) on the task (e.g., recognizing the disease).
    \item Given two different explanations from different algorithms, which one provides a better quality explanation? This is also known as ``Binary Forced Choice'' evaluation metric \cite{doshi2017towards}. This test can be used to compare the different explanations from different interpretable models.
\end{itemize}
\textbf{Non-human Based Evaluation Metrics.} Multiple factors such as human fatigue, improper practice sessions and incentive costs can affect experimental results when human-subject evaluation metrics are used. Hence, it is important to conduct other evaluation metrics.
 %The proposed model on interpretability should be compared with the original model to make sure that they are faithful to it. Below are the common approaches to this mean:
\begin{itemize}
    %\item \mk{An interpretable base classifier introduced in \ref{inr_clssi} is used as a proxy model. This will be used to measure how much the outcome of the model matches the outcome of the proxy model.}
    \item How much a proposed interpretable model recovers the important features of the data for a certain prediction task?
    This requires the important features to be known beforehand. We should verify that the model will pick up the important features of the data. One simply can use any base method introduced in section \ref{inr_clssi} as a proxy model to extract the important features. The fraction of these important features recovered by the interpretable method can be used as an evaluation score \cite{ribeiro2016should}.
    \item How locally faithful the proposed method is compared to the original model (fidelity)? Lack of fidelity will result in a limited insight to the original model \cite{yang2019evaluating}. In convolutional neural network, one common approach is the image occlusion. The pixels that the interpretable method defines as important will be masked to see whether it reflects on the classification score or not \cite{selvaraju2017grad, zeiler2014visualizing}.
    \item How consistent the explanations are for the similar instances with the same class label? The explanations should not be significantly different for samples with the same label with a slightly different features. This instability could be the result of a high variance as well as the non-deterministic components of the explanation method \cite{molnar2019interpretable}.
    
\end{itemize}
\subsubsection{Causal Evaluation Metrics}

 % \rg{If possible, could you show two examples here? Let one be causal and the other one be not causal so the readers can understand what a causal explanation means after reading these two examples.} \mk{No need for the examples that Ruocheng requested, this is almost the end of the paper and the readers already know about what is causal or not.}
%
%\rg{The evaluation metrics of causal interpretability have to cover every type of causal interpretability methods discussed in Section 3.3. If there is no such a method for a certain type (e.g., Section 3.3.3), we have to point it out (better with reasons why the evaluation of such methods is difficult, what do we need to create such a evaluation method). I remember you mentioned there are methods for causal inference and fairness (Section 3.3.4).}
Due to the lack of groundtruth for causal explanations, to verify the causal aspect of the proposed framework, we need to quantify the desired characteristics of the model and measure the ``goodness'' of them via some predefined proxy metrics.
In the following, we go over the existing metrics to evaluate the proposed causal interpretable frameworks for different categories of causal interpretability.

\textbf{Counterfactual Explanations Evaluation Metrics.}
%\textbf{properties of a good counterfactual explanation}:
Existing approaches for causal interpretability are mostly based on generating counterfactual explanations. For such approaches, the causal interpretability is often measured through the goodness of generated counterfactual explanation.
As mentioned in section \ref{causalinterpret}, a counterfactual explanation is the highest level of explanation and therefore, we can claim that if an explanation is a counterfactual explanation and is generated by considering causal relationships, it is indeed explainable. However, due to the lack of groundtruth for counterfactuals, we are unable to measure if the generated explanations are generated based on causal relationships. Therefore, to measure the ``goodness'' of counterfactual explanations, we suggest to conduct experiments to (1) measure the interpretability of the explanations using the metrics designed for interpretability; and (2) evaluate the conterfactuals themselves by measuring different characteristics of them. An interpretable Counterfatual explanation should have the following characteristics: 
\begin{itemize}
    
    \item The model prediction on the counterfactual sample ($x_{cf}$) needs to be close to the predefined output for counterfactual explanation.
    
    \item  The perturbation $\delta$ changing the original instance x into
    $x_{cf} = x + \delta$ should be sparse. In other words, size of counterfactual (i.e., number of features) should be small.
\item A counterfactual explanation $x_{cf}$ is considered interpretable if it lies close to the model’s training data distribution. %\raha{one paper GDPR claims different!}
\item The counterfactual instance $x_{cf}$ needs to be found fast enough to ensure it can be used in a real life setting.
%\item feasibility of the counterfactual actions given user context and constraints (some features such as age and race can not be changed) explanations should be actionable
\item  Counterfatual explanations generated for a data instance should be different from each other. In other words, counterfactual explanations should be diverse.
\item Visual-linguistic counterfactual explanations must satisfy the following two criteria, (1) Visual explanation is the region which keeps high
positiveness/negativeness on the model prediction for specific positive/negative classes; (2) Linguistic explanation should be compatible to the visual counterpart in the generated visual explanations. 
    
\end{itemize}

\begin{table*}
\begin{tabular}{|c|c|c|l|}
 \hline
 \multicolumn{4}{|c|}{Overview of interpretable models and their categories} \\
 \hline
\multirow{4}{6.5em}{Traditional Interpretability} & \multirow{2}{16em}{Interpretable Models: \cite{yang2016hierarchical}, \cite{xu2015show}, \cite{lu2017knowing}, \cite{xu2016ask}, \cite{lu2016hierarchical}, \cite{yang2016stacked}, \cite{kingma2013auto}, \cite{chen2016infogan}, \cite{higgins2017beta}} & \multirow{4}{6.5em}{Causal Interpretability}& Model-based: \cite{DBLP:journals/corr/abs-1811-04376}, \cite{DBLP:journals/corr/abs-1802-00541}, \cite{DBLP:journals/corr/abs-1902-02302}, \cite{DBLP:journals/corr/abs-1812-03253}, \cite{zhao2019causal}, \cite{parafita2019explaining}, \cite{DBLP:journals/corr/abs-1811-10597},\\
 \cline{4-4}
 &  &  & \vtop{\hbox{\strut Example-based: \cite{DBLP:journals/corr/abs-1811-05245}, \cite{DBLP:journals/corr/abs-1812-01263}, \cite{DBLP:journals/corr/abs-1806-09809}, \cite{DBLP:journals/corr/abs-1711-00399}, \cite{DBLP:journals/corr/abs-1905-07697}, \cite{DBLP:journals/corr/abs-1906-09293},}\hbox{\strut \cite{DBLP:journals/corr/abs-1906-10671}, \cite{DBLP:journals/corr/abs-1907-02584}, \cite{DBLP:journals/corr/abs-1907-03077}, \cite{DBLP:journals/corr/abs-1907-07165},
 \cite{DBLP:journals/corr/abs-1904-07451}}} \\
 \cline{2-2} \cline{4-4}
 & \multirow{2}{16em}{Post-hoc: \cite{kim2016examples}, \cite{ribeiro2016should}, \cite{lundberg2017unified}, \cite{simonyan2013deep}, \cite{selvaraju2017grad}, \cite{zeiler2014visualizing}, \cite{erhan2009visualizing}, \cite{craven1996extracting}, \cite{towell1993extracting}, \cite{lou2012intelligible}} &  & Fairness: \cite{DBLP:conf/nips/KusnerLRS17}, \cite{NIPS2017_6668}, \cite{DBLP:journals/corr/abs-1809-02519}, \cite{inproceedings}\\
 \cline{4-4}
 &  &  & Guarantee: \cite{DBLP:journals/corr/abs-1901-08576}, \cite{doshi2017towards}, \cite{ribeiro2016model}\\
\hline
\end{tabular}
\caption{A summary of the state-of-the-art frameworks for each type of interpretability}
\label{table:overview}
\end{table*}
%\textbf{}More specifically, most existing approaches in this area are based on generating counterfactual explanations.
%
%\rg{Maybe you want to add a sentence to connect the causal interpretability of a model and the goodness of a counterfactual explanation. For example, for such approaches, the causal interpretability is often measured through the goodness of generated counterfactual explanation.}
%
Below, we briefly discuss these evaluation metrics designed to assess aformentioned characteristics of a counterfactual explanation:

%\raha{Sparsity and Size}

To evaluate the sparsity of the generated counterfactual examples, Mc Grath et al. \cite{DBLP:journals/corr/abs-1811-05245} measures the size of a generated example by counting the number of features each example consists of. Van Looveren and Klaise \cite{DBLP:journals/corr/abs-1907-02584} use elastic net loss term $EN(\delta) = \beta ||\delta||_1 + ||\delta||_2^2$ where $\delta$ is the distance between the original instance and its generated counterfactual example and $\beta$ is the hyperparameter.

%counterfactuals with small number of features are more interpretable, 
%Mc Grath et al. claim that counterfactual explanations with smaller number of features are more interpretable to humans . Therefore, in order to measure if the generated counterfactual explanation is good, they proposed to measure the size of the explanations by counting the number of features each of them consists of.

%One metric to measure how good a counterfactual example is, is the size of counterfactual explanations. Counterfactual explanations are encouraged to have smaller size, i.e. consist of lesser number of features. In order to demonstrate that generated counterfactuals are "good", one way is to measure the size of generated counterfactuals \cite{DBLP:journals/corr/abs-1811-05245} \rg{This sentence is redundant.}.
%Another way to evaluate the generated counterfactuals is to evaluate them based on their interpretability, sparsity and speed of the search process. The sparsity is evaluated using the elastic net loss term $EN(\delta) = \beta. ||\delta||_1 + ||\delta||_2^2$ \rg{The symbols, $\beta$ and $\delta$, need to be explained.}

%while the speed is measured by the time and the number of
%gradient updates required until a satisfactory counterfactual \rg{example?}
%$x_{cf}$ is found \cite{DBLP:journals/corr/abs-1907-02584} \rg{we need to write a sentence to connect counterfactual explanations to counterfactual examples.}.
%

%\raha{Interpretability of the counterfactual}

In order for counterfactual explanations to be interpretable, they need to be close to the  data manifold. Looveren and Klaise improves this criterion by suggesting that the counterfactuals are interpretable if they are close to the data manifold of the counterfactual class \cite{DBLP:journals/corr/abs-1907-02584}.  
To measure the interpretability defined above,  Looveren and Klaise propose to measure the ratio of the reconstruction errors when the model used for generating counterfactuals is trained only on the counterfactual class vs when it is trained on the original class \cite{DBLP:journals/corr/abs-1907-02584}. The proposed metric is shown in the following equation,
\begin{equation}
\label{eq10}
IM1(AE_i, AE_{t_0}, x_{cf}) = \frac{||x_0+\delta - AE_i(x_0+\delta) ||_2^2}{||x_0+\delta - AE_{t_0}(x_0+\delta) ||_2^2 +\epsilon}
\end{equation}
 Where $AE_i$ and  $AE_{t_0}$ represent the autoencoders used to generate the counterfacutals trained on the class $i$ (counterfactual class) and class $t_0$ (the original class), respectively. We let $x_{cf}$ and $x_0$ be the counterfactual explanation and the original sample. In addition, $\delta$ denotes the distance between the original and counterfactual samples.
 A lower value of $IM1$ shows that counterfactual examples can be better reconstructed from the autoencoder trained on the counterfactual class in comparison to the autoencoder trained on the original class. This implies that the generated counterfactuals are closer to the counterfactual class data manifold.

Another metric proposed by \cite{DBLP:journals/corr/abs-1907-02584} measures how similar the generated counterfactuals are when generated using the autoencoder trained on only counterfactuals vs the autoencoder trained on all classes. The metric is shown in the following equation,

\begin{equation}
\label{eq11}
IM2(AE_i, AE_{t_0}, x_{cf}) = \frac{||AE_i(x_0+\delta) - AE(x_0+\delta) ||_2^2}{||x_0+\delta||_1 +\epsilon}
\end{equation}

A lower value of $IM2$ shows that counterfactuals generated by both autoencoders trained on all classes and counterfactuals are more similar. This implies that the generated counterfactual distribution is as good as the distribution over all classes.
%\cite{DBLP:journals/corr/abs-1907-02584} proposes to measure the ratio which shows how well the model can reconstruct the counterfactuals using the autoencoder trained on only counterfactuals versus authoencoder trained on the original data \rg{If possible, please provide the formal definitions of all the metrics mentioned in this subsection.}. This metric gauges how close the counterfactual examples are to the data manifold.

% \raha{validity}

%\raha{Understanding of local decision boundary by human} 

Generated counterfactual explanations can be used to measure users' understanding of a machine learning model's local decision boundary.
Mothilal et al. \cite{DBLP:journals/corr/abs-1905-07697} propose to mimic users' understanding of a model's local decision boundaries by, (a) constructing an auxiliary classifier on both original inputs and counterfactual examples; and (b) measuring how well it mimics the actual decision boundaries. More specifically, they train a 1-nearest neighbor (1-NN) classifier on both the original and the counterfactual samples to predict the class of new inputs. The accuracy of this model is then compared with the accuracy of the original model.
%\mk{This is vague. What is the auxiliary model? does it need to be similar to the first model?}

%\raha{Proximity}

The definition of counterfactual explanations implies that generated explanations should be as similar as possible to the original instance.
In order to evaluate the proximity between original samples and counterfactual explanations, Mothilal et al. \cite{DBLP:journals/corr/abs-1905-07697} defines proximity as Eq. \eqref{eq17},
\begin{equation}
\label{eq17}
\begin{split}
& Proximity = -\frac{1}{k} \sum_{i=1}^{k} dist(x_{cf_i}, x)\\
\end{split}
\end{equation}
In order to be able to calculate the proximity for both categorical and continuous features, the authors further propose two metrics to calculate the proximity for categorical and continuous features. For continuous features, the proximity is defined as the mean of feature-wise $L_1$ distances between the original sample and counterfactuals divided by the median absolute deviation (MAD) of the feature’s values in the training set. For categorical features, disctance function is calculated such that for each categorical feature it assigns 1 if the feature differs from the original feature and otherwise it assigns 0. %Equation \eqref{eq13} shows the mathematical definition of proximity for continuous features:
 
% \begin{equation}
%\label{eq13}
%\begin{split}
%& continuous\_proximity = -\frac{1}{k} \sum_{i=1}^{k} continuous\_dist(c_i, x)\\
%& \text{ Where ~~} continuous\_dist(c, x) = \frac{1}{d} \sum_{p=1}^{d} \frac{||c_p - x_p||}{MAD_p}
%\end{split}
%\end{equation}

%Where d is the number of continuous variables.

%To calculate the proximity between 
%\raha{Speed}

% measutring the speed
In order to gauge the speed of generating counterfactual explanations, Looveren and Klaise \cite{DBLP:journals/corr/abs-1907-02584} measure the time and the number of gradient updates until the desired counterfactual explanation is generated.

%\raha{diversity}

Diversity of generated counterfactuals is measured via measuring feature-wise distances between each pair of counterfactual examples and calculating diversity as the mean of the distances between each
pair of examples \cite{DBLP:journals/corr/abs-1905-07697}. Eq. \eqref{eq12} illustrates the measure used for diversity.
\begin{equation}
\label{eq12}
Diversity = \frac{1}{|C_k|^2} \sum_{i=1}^{k-1} \sum_{j=i+1}^k d(x_{cf_i}, x_{cf_j})
\end{equation}

Where $C_k$ represents a set of k counterfactuals generated for the original input, $x_{cf_i}$ and $x_{cf_j}$ are the i-th and j-th counterfactuals in the set $C_k$.
%
%\rg{Notation issues in Eq. (12): 1. what is the square of a set? (In addition, subscript is often used to mean the $k$-th component/instance of a set/matrix/vector instead of its size) 2. although it can be implied, we have to point out how is the distance function $d(\cdot,\cdot)$ defined. 3. it seems $c_i$ and $c_j$ are not defined, even I know they are the $i$-th and $j$-th counterfactual examples, we have to literally say it. This also raises a consistency issue where $x_{cf}$ was used to denote a counterfactual example.}

%\raha{visual-linguistic counterfactual explanations}

Kanehira et al. \cite{DBLP:journals/corr/abs-1812-01263} propose metrics to evaluate visual-linguistic counterfactual explanations to ensure, (a) visual explanations keep possession of high  positiveness/negativeness on the model predictions for
positive/negative classes; (b) linguistic explanations are compatible with their corresponding visual explanations.
%
%They claim that visual-linguistic counterfactual explanations must satisfy the following two criteria: (1) Visual explanation is the region which retains high
%positiveness/negativeness on the model prediction for
%specific positive/negative classes. (2) Linguistic explanation is compatible to the visual counterpart. 
To measure if the generated examples meet these criteria, authors in \cite{DBLP:journals/corr/abs-1812-01263} propose two metrics based on the accuracy.
More specifically, to check for the first condition, they
investigate how the output of the target classifier changes
towards the negative class when a specific region is removed from the input. To measure the second criterion, for each output pair (s, R) they examine how the region R makes the concept s distinguishable by humans. To measure this quantitatively, they compute the accuracy by utilizing bounding boxes for each attribute in the test set. More specifically, IoU (intersection over union) between a
given R and all bounding boxes $R_0$ corresponding to
attribute $s_0$ is calculated. Then the accuracy is measured by selecting the the attribute $s_0$ with the largest IoU score and checking its consistency with s a counterpart of R. 

Table \ref{table:counteval} summarizes evaluation metrics for counterfactulas explanations based on the properties of the generated examples.

%\raha{axioms}

\textbf{Model-based Evaluation Metrics.}
Due to the lack of evaluation groundtruth for representing the actual effect of each component of the model on its final decisions, evaluation for this type of models is still an open problem. One common way of evaluating such models is to report the most important components of a model by measuring their causal effects on the outcome of the model \cite{DBLP:journals/corr/abs-1802-00541, DBLP:journals/corr/abs-1811-04376}. Chattopadhyay et al. also used the causal attribution of each neuron on the output %($ACE^y_{do(\alpha)} ($) 
%\rg{this notation of ACE is not consistent with the definition of ACE in Section 3. The usage of superscript (e.g., $y$) and subscript (e.g., $do(\alpha)$) are not defined, in addition, $\alpha$ itself is not defined. For notations, I would like to mention two standards for you to check if any equation/notation in a work is clear: (1). after reading an equation/notation, is it self explained or at least within the next few sentences, every symbol used in it has been explained. (2). The usage of notations has to be consistent, which means any reference to the same symbol has the same meaning. That's why maintaining a table of notations can help.}
to visualize the local decisions of the model by saliency map. Moreover, to further investigate how well the model estimates the ACE, they proposed to run the model on datasets for causal effect estimations \cite{DBLP:journals/corr/abs-1902-02302}.
%\rg{Maybe you want to add a table to summarize the metrics mentioned in this subsection.}

\textbf{Causal Fairness Evaluation.}
Evaluation of causal fairness models is a challenging task. Papers in this field usually assess the performance of the model for detecting discrimination. Zhang et al. leverage direct, indirect and spurious effect measures (defined in section \ref{fairness}) to detect and explain discrimination \cite{inproceedings}. However, to the best of our knowledge, no quantitative measure of causality of a fairness algorithm existis.

%problems with the exisiting methods for causal interpretation
%\section{Related areas}
\section{Conclusion}
In this survey, we introduce the problem of interpretability in machine learning. We view the problem from two perspectives, (1) Traditional interpretability algorithms; (2) causal interpretability algorithms.  However, the primary focus of the survey is on causal frameworks. We first provide different definitions of interpretability, then review the state-of-the-art methods in both categories and point out the differences between them. Each type of interpretable models is further subdivided into other sub categories to provide readers with better overview of existing directions and approaches in the field. More conceretely, for traditional methods, we divide existing work into inherently interpretable models and post-hoc intrerpretability. For causal models, we divide the existing works into the following four categories: counterfactual examples, model-based interpretability, causal models in fairness and interpretability for verifying causal relationships.
We also address the challenging problem of evaluating interpretable models , explain existing metrics in detail and categorize them based on the scenarios they are designed for. Table \ref{table:overview} summarizes state-of-the-art methods which belong to each category of interpretability.
%Recently, causal interpretable models have gained a lot of attentions. In this survey, we give a comprehensive overview of definition of causal-interpretbility, state-of-the-art methods in this field and commoly used evaluation metrics and different scenarios they are designed for. 

\section*{ACKNOWLEDGEMENTS}
We would like to thank Andre Harrison for helpful comments.

%
% The next two lines define the bibliography style to be used, and the bibliography file.
\bibliographystyle{abbrv}
\bibliography{sigkddExp}

\clearpage
% 
% If your work has an appendix, this is the place to put it.
%\appendix
%\section{Appendix.}
%\subsection{Table of Symbols}
%Table \ref{table:note} shows the notations used in this paper.
%\begin{table*}
%\begin{tabular}{|c|c|}
% \hline
%\textbf{Notation} & \textbf{Description}  \\
% \hline
% x& Feature of an input or component of a model for which we would like to estimate the causal effect on the final decision of the model\\
%\hline
%y& Final decision made by a model/ Output of the model\\
%\hline
%$x_{cf}$& Counterfactual explanation for input x\\
%\hline
%$y_{cf}$& Counterfactual outcome for input x when x is set to $x_{cf}$\\
%\hline
%$\delta$& Perturbations added to $x$ to generate $x_{cf}$ ($x_{cf} = x + \delta$)\\
%\hline
%$do (x = \alpha)$& Intervention on variable $x$ and set it to $\alpha$\\
%\hline
%CBN & Causal Bayesian Netwrok\\
%\hline
%SEM/SCM & Structural Equation Model/ Structural Causal Model\\
%\hline
%ACE & Average Causal Effect ($ACE = %\EX[y|do(x = 1)] - \EX[y|do(x = 0)]$) \\
%\hline
%$IE_{x_0, x_1} (y|x)$ & Indirect Effect %($IE_{x_0, x_1} (y|x) = P(y_{x_0,W_{x1}} %|x) - P(y_{x_0} |x)$)\\
%\hline
%$DE_{x_0, x_1} (y|x)$ & Direct Effect %($DE_{x_0, x_1} (y|x) = P(y_{x_1,W_{x0}} %|x) - P(y_{x_0} |x)$)\\
%\hline
%$SE_{x_0, x_1} (y|x)$ & Spurious Effect 
%($SE_{x_0, x_1} (y|x) = P(y_{x_0} |x_1) - %P(y|x_0))$\\
%\hline
%\end{tabular}
%\caption{Table of symbols}
%\label{table:note}
%\end{table*}

\end{document}